\documentclass[sn-mathphys-num]{sn-jnl}

\usepackage{graphicx}
\usepackage{multirow}
\usepackage{amsmath,amssymb,amsfonts}
\usepackage{amsthm}
\usepackage{mathrsfs}
\usepackage[title]{appendix}
\usepackage{xcolor}
\usepackage{textcomp}
\usepackage{manyfoot}
\usepackage{booktabs}
\usepackage{algorithm}
\usepackage{algorithmicx}
\usepackage{algpseudocode}
\usepackage{listings}

\usepackage[most]{tcolorbox}

\usepackage{color}
\usepackage{soul}
\definecolor{lightyellow}{rgb}{0.97, 0.97, 0}
\sethlcolor{lightyellow}

\usepackage{placeins}

\begin{document}

	\title{PVLens: Enhancing Pharmacovigilance Through Automated Label Extraction}
	
	
	\author[1]{\fnm{Jeffery L.} \sur{Painter}}\email{jeffery.l.painter@gsk.com}
	\author[1]{\fnm{Greg} \sur{Powell}}\email{gregory.e.powell@gsk.com}
	\author[2,3]{\fnm{Andrew} \sur{Bate}}\email{andrew.x.bate@gsk.com}

	\affil*[1]{\orgname{GlaxoSmithKline}, \orgaddress{\city{Durham}, \state{NC}, \country{USA}}}
	\affil[2]{\orgname{GlaxoSmithKline}, \orgaddress{\city{London}, \country{UK}}}
	\affil[3]{\orgname{London School of Hygiene and Tropical Medicine}, \orgaddress{\city{London}, \country{UK}}}

	
	\abstract{Reliable drug safety reference databases are essential for pharmacovigilance, yet existing resources like SIDER are outdated and static. We introduce PVLens, an automated system that extracts labeled safety information from FDA Structured Product Labels (SPLs) and maps terms to MedDRA. PVLens integrates automation with expert oversight through a web-based review tool. In validation against 97 drug labels, PVLens achieved an F1 score of 0.882, with high recall (0.983) and moderate precision (0.799). By offering a scalable, more accurate and continuously updated alternative to SIDER, PVLens enhances real-time pharamcovigilance with improved accuracy and contemporaneous insights.}

	\keywords{Pharmacovigilance, Natural Language Processing (NLP), Drug Safety, ADR}
	
	\maketitle

	\section{Introduction}
		
		A clear understanding of known adverse effects, along with continuous surveillance for emerging safety concerns, is essential for patients, healthcare professionals, and pharmacovigilance (PV) scientists. Structured reference sets, such as FDA-approved drug labels, are critical for safety evaluation. One of the most widely used reference sets in PV is the FDA-approved drug label, which provides official safety and efficacy information for medicinal products in the US.
		
		Despite their importance, no continuously updated, gold-standard resource exists for systematically accessing labeled drug events \cite{smith2013lessons}. The Side Effect Resource (SIDER) has been widely used in drug discovery and PV workflows \cite{kuhn2016sider}, yet it has not been updated since 2015, making it increasingly misaligned with current safety knowledge.  Updates in MedDRA terminology have led to obsolete, incorrect, and misleading mappings, affecting drug safety assessments —for example, SIDER lists “urine output” as an adverse event, yet it was reclassified as a quantitative concept in the UMLS, falling outside SIDER's own original inclusion criteria. At least 40 similar terms have been reclassified. Additionally, SIDER’s reliance on PubChem IDs has introduced significant annotation errors, such as misclassifying Lescol with CID-124838623—an identifier now assigned to Remdesivir. These limitations undermine its reliability, particularly for newer drugs requiring rigorous safety monitoring.
		
		Recent efforts have explored machine learning (ML)-based extraction methods, from BERT-based models to LLM-driven systems like OnSIDES and AskFDALabel \cite{tanaka2024onsides, wu2025leveraging}. While scalable, these approaches currently struggle with MedDRA-specific mappings and hierarchical relationships \cite{dong2024optimizing}. For example, OnSIDES mapped 4,423 distinct MedDRA terms to adverse events and indications, whereas our pipeline mapped over 8,640 distinct terms, demonstrating a limitation in LLM abilities to robustly capture adverse event terminology. More broadly, these ML-based and term-matching systems attempt to infer product mappings without leveraging structured biomedical resources, such as UMLS RxNorm and MTHSPL (Metathesaurus SPL entries), introducing redundant complexity into the extraction process.
		
		Rather than relying on ML-based substance identification, PVLens directly integrates UMLS and RxNorm mappings, eliminating the need for inferred mappings. Since RxNorm is derived from SPL data, it provides the most reliable linkage between drug substances and labeled adverse events. This structured, verifiable approach enhances accuracy, reduces ambiguity, and improves scalability for PV analytics.
		
		To address these challenges, we introduce PVLens, an automated system that extracts safety data from FDA Structured Product Labels (SPLs) and maps terms to MedDRA, RxNorm, and SNOMED CT. PVLens processes SPL XML data using dictionary-based NLP and UMLS resources \cite{bodenreider2004unified}, while integrating Safety-Related Labeling Changes (SrLC) \cite{munesh2024regulatory}. Optimized for large-scale processing, PVLens can extract and process over 50,000 SPLs in under an hour, ensuring frequent updates aligned with evolving safety information.  While some commercial pharmacovigilance tools exist, they are proprietary and lack transparency, with evaluations revealing incomplete substance mappings. In contrast, PVLens will be released as an open-source resource, ensuring accessibility and reproducibility for PV analytics.
		
		To assess its effectiveness, we conducted a structured validation study comparing PVLens’ automated extractions to expert-reviewed annotations across 97 drug labels (n=63 post-2014 with documented labeled event changes, n=34 pre-2014)\footnote{Of the 250 labels initially selected for review, 97 have been completed and reported here. The review is ongoing, and the full data set will be made available once this process is finished.}. This study evaluates how well PVLens captures labeled safety information and aligns with expert judgment. Unlike SIDER, which has been widely cited but never formally validated, we believe PVLens provides the first benchmarked system for structured label extraction. Additionally, as no officially reviewed MedDRA mappings exist for post-2014 products, this evaluation represents the first structured comparison of extracted label content using a reproducible methodology.

	\section{Methods}
	
		\subsection{System Overview and Data Extraction}
		
\begin{figure}[htbp]
				\centering
				\fbox{\includegraphics[width=0.85\textwidth]{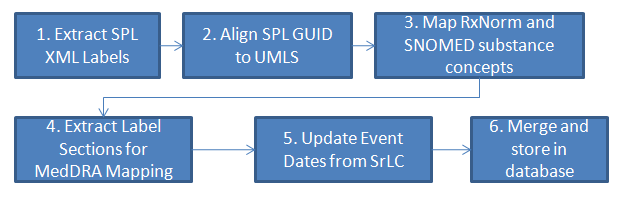}}
				\caption{PVLens Processing Pipeline Overview. [GUID = Global Unique Identifier, SrLC = Safety-Related Label Change]}
				\label{fig:fig_01}
			\end{figure}

			PVLens employs a multi-phase extraction pipeline to ensure precise term identification and MedDRA mapping, as outlined in Figure \ref{fig:fig_01}. The process begins with a custom XML parser extracting the labeled sections of interest (i.e. indications, adverse events, black box warning) from SPLs (Step 1). Each label is then mapped to the UMLS using its Global Unique Identifier (GUID) by way of MTHSPL (Step 2), followed by RxNorm and SNOMED mappings to NDC codes (Step 3). Extracted text is mapped to MedDRA terms using NLP-based pattern matching (Step 4), and reported safety event dates are updated using SrLC (Step 5). The final outputs are merged into the PVLens repository, ensuring continuous updates for pharmacovigilance (Step 6).
				
			The entire pipeline processes all prescription SPLs in under an hour, covering 5,358 distinct substances. To reduce noise in MedDRA mapping, generic phrases (e.g., adverse reaction) are filtered using a predefined stop word list\footnote{\url{https://gist.github.com/sebleier/554280}}, refined based on prior work \cite{demner2018dataset}.	

		\subsection{Study Design and Product Selection}
		
			To evaluate PVLens, we conducted a structured validation using independent review and adjudication of extracted terms. 97 distinct substance SPL labels were selected for review. Of these, 63 were approved post-2014 and had a labeled event change or black box warning added within two years post-approval, ensuring relevance for evaluating PVLens’ ability to assist in the detection of emerging safety signals. The remaining 34 labels were selected at random and contained at least one mapped indication and one adverse event.

		\subsection{Review and Adjudication Process}
		
			A total of 12 reviewers, including second- to fourth-year pharmacy students and a senior pharmacovigilance scientist, participated in the study. Each label was independently reviewed by two randomly assigned reviewers, with discrepancies on MedDRA term inclusion resolved by an expert adjudicator (a PharmD, PV scientist with 20+ years of experience). The 97-label review and adjudication process were completed within three weeks, requiring approximately 57 hours (excluding adjudication). The mean review time per label was 17.2 minutes, with a median of 28 minutes. Some complex labels required over 100 minutes for review due to their detailed safety content.
			
			Beyond discrepancy resolution, reviewers could propose additional terms absent from the NLP-extracted outputs. These user-added terms were consolidated and reassessed by the adjudicator for final validation.

	\section{Results}	
			
		Reviewers were assigned labels based on availability, meaning that the study did not follow a paired review design. Consequently, traditional inter-rater reliability (IRR) metrics, such as Cohen’s kappa, were not applicable since reviewers did not consistently evaluate the same set of labels. Instead, overall agreement across the 97 labels was measured, with independent reviewers agreeing on MedDRA term inclusion or exclusion 77\% of the time.  Adjudicator-reviewer agreement was consistently high, with a median of 91.3\%, reinforcing the reliability of the review process. The limited number of false negatives (FN = 79, 1.3\% of total terms reviewed) further supports the internal consistency of the review process and indicates that the vast majority of MedDRA term assignments were confirmed across multiple reviewers.
		
		The algorithm was most effective in detecting adverse events, followed by indications, with black box warnings presenting the most challenges (Table \ref{tab:performance}). High recall across all categories suggests minimal missed MedDRA terms, though precision varied, indicating a tendency to over-include terms that required expert adjudication.
		
		\begin{table}[h]
			\centering
			\begin{tabular}{lcccccc}
				\toprule
				\textbf{Category} & TP & FP & FN & \textbf{Precision} & \textbf{Recall} & \textbf{F1-score} \\
				\midrule
				Adverse Event & 4,223 & 887 & 66 & 82.6\% & 98.5\% & 89.9\% \\
				Indication & 170 & 110 & 10 & 60.7\% & 94.4\% & 73.9\% \\				
				Black Box Warning & 187 & 155 & 3 & 54.7\% & 98.4\% & 70.3\% \\
				\midrule
				\textbf{Overall} & 4,580 & 1,152 & 79  & \textbf{79.9\%} & \textbf{98.3\%} & \textbf{88.2\%} \\
				\bottomrule
			\end{tabular}
			\caption{Validation study results}
			\label{tab:performance}
		\end{table}
		
		\subsection{False Negative Analysis}
		To assess false negatives, reviewers could manually select or add terms to each section under review. These sets of terms were then mapped to MedDRA when possible, for our false negative analysis, and was classified as a false negative only if it met three criteria: (1) it belonged to a valid semantic type, (2) it was not already mapped to a MedDRA synonym, and (3) it was not excluded as a stop word by PVLens.
		
		Of 1,012 user-added terms, only 79 (7.8\%) were successfully mapped to MedDRA Preferred Terms (PT) or Lower Level Terms (LLT), comprising 3 black box warnings, 10 indications, and 66 adverse events. Most excluded false negatives were due to synonym recognition or classification differences, with 12 terms already captured by MedDRA synonyms and 12 classified as outside valid semantic types. Additionally, 9 terms were excluded based on predefined stopword filters, ensuring only clinically relevant terms were retained. The low false-negative rate suggests PVLens effectively captures relevant terms, with omissions primarily due to synonym variations or nuanced phrasing.	
	
	\section{Discussion}	
			
		PVLens serves as a continuously updated, transparent alternative to static resources like SIDER, enabling real-time tracking of evolving drug safety label information. By incorporating temporal tracking and the SrLC, PVLens enhances drug safety relevance. Future work will focus on improving precision, potentially by integrating LLMs for automated adjudication and refining term-matching approaches. 
		 
		PVLens prioritizes high recall to ensure comprehensive capture of labeled adverse events (AEs), as missing critical AEs poses a greater risk than including extraneous terms. While recall exceeded 98\%, precision was 80\%, indicating that the algorithm captures more MedDRA PTs than required. Reviewer input confirmed only 79 terms were missed (3 black box warnings, 10 indications, 66 AEs), reinforcing strong MedDRA coverage.
		
		For routine PV, these findings suggest that PVLens is a valuable resource requiring minimal further refinement. Its high recall ensures comprehensive capture of labeled AEs and indications, reducing the risk of missing critical safety information. While expert review remains necessary, PVLens provides a structured, scalable repository for US-approved products, supporting both routine screening and structured case evaluation.
		
		PVLens minimizes false negatives (FNs), reducing the likelihood of missing critical AEs. While false positives (FPs) occur, they are manageable through rapid adjudication, a far more efficient process than manually reviewing every SPL. This significantly improves efficiency over traditional manual label reviews, which are time-consuming and inconsistently applied. Our performance metrics align with MITRE and FDA assessments of NLP techniques for AE extraction, which reported up to 79\% F1 for MedDRA coding, compared to PVLens' 88.2\% \cite{bayer2021ade}. Similarly, NLP-based MedDRA annotation of drug labels have shown F1 scores ranging from 67\%-79\%, highlighting the inherent challenges of optimizing extraction performance \cite{ly2018evaluation}.
		
		Other initiatives, such as RS-ADR \cite{lee2022data}, integrate EHR and real-world data look to validate adverse drug reaction (ADR) signals, enhancing post-market surveillance. However, such approaches rely on retrospective clinical data, which is subject to reporting bias and coding inconsistencies. PVLens complements these efforts by offering a structured, up-to-date resource, ensuring a comprehensive repository of AEs and indications. Integrating structured label extractions with real-world validation can further strengthen PV.
		
		Advancements in biomedical informatics offer new opportunities to refine PVLens beyond rule-based extractions. Future iterations will explore context-aware embeddings to improve term-matching and reduce FPs. Additionally, integrating semantic reasoning with UMLS, MedDRA, and RxNorm could enhance concept linkage across regulatory datasets.
		
		Harmonizing safety information across global regulatory bodies remains a challenge. PVLens’ ontology-driven approach enables expansion beyond FDA SPLs to EMA, PMDA, and WHO reports, fostering global PV alignment. To our knowledge, PVLens is the first validated comprehensive ADR database, systematically evaluated against human reviewers. By combining structured label extraction with AI-driven surveillance, PVLens lays the foundation for hybrid PV systems, integrating regulatory knowledge with real-world evidence. As agencies adopt AI-driven monitoring, PVLens offers a scalable, interpretable, and auditable framework for drug safety analytics.

	\section{Conclusion}

		Modern pharmacovigilance requires accurate, continuously updated, and validated resources. PVLens meets this need by extracting and mapping labeled safety events from FDA SPLs to MedDRA, RxNorm, and SNOMED. With high recall and solid precision, achieved through dictionary-based tokenization and context-aware filtering, PVLens enhances pharmacovigilance workflows and supports structured safety assessments.
		
		Validation against human reviewers confirms PVLens’ ability to comprehensively capture labeled safety information, making it well-suited for routine screening and case evaluation. While improving precision remains a priority, high recall minimizes the risk of missing critical safety events. Future work will focus on expanding beyond U.S. products and integrating LLMs and AI-driven techniques to refine term matching.
		
		To our knowledge, PVLens is the first validated ADR database, systematically evaluated against human reviewers. By offering a continuously updated, reproducible repository of labeled indications and AEs, PVLens provides a scalable, evolving alternative. It strengthens the foundation for drug safety analytics, enhancing risk assessment and signal evaluation in real time.

	\section{Declarations}
	
		GSK covered all costs associated with the conduct of the study and the development of the manuscript and the decision to publish the manuscript. J.P., G.P, and A.B. are employed by GSK and hold financial equities. This manuscript has not been submitted to, nor is under review at, another journal or other publishing venue. The authors have no competing interests to declare that are relevant to the content of this article.

		\textbf{Author Contributions} JLP and AB contributed to	the study concept, data acquisition, data analysis, and data interpretation. GP contributed to data interpretation. \\
		
		\textbf{Data availability} All data and code to support this analysis is available from \url{https://github.com/jlpainter/AMIA2025/tree/main/pvlens}.\\		
	
		\section{Acknowledgments}
			
			We acknowledge the UNC Eshelman School of Pharmacy PharmD students for reviewing SPL data under supervision of Greg Powell, PharmD: Miranda Barker, Hibbah Ashraf, Megan Earnhart, Margaux Meilhac, Breannah Keys, Ravi Parekh, Emily Wu, Vicky Mei, Derek Bassett, Joshua Airgood, Ryan Le and Rowena Dzorvakpor.

	\bibliography{spl}

\end{document}